\documentclass[letterpaper, 10 pt, conference]{ieeeconf}  

\IEEEoverridecommandlockouts                              

\usepackage{graphicx}
\usepackage{amsmath} 
\usepackage{amssymb}  
\usepackage[utf8]{inputenc}
\usepackage{balance}
\usepackage[detect-all]{siunitx}
\usepackage{dblfloatfix}
\usepackage{times}
\usepackage{epsfig}
\usepackage{booktabs}
\usepackage{placeins}
\usepackage{todonotes}
\newcommand{\etal}{\textit{et al}. }

\title{\LARGE \bf
PST900: RGB-Thermal Calibration, Dataset and Segmentation Network 
}

\author{Shreyas S. Shivakumar, Neil Rodrigues, Alex Zhou, Ian D. Miller, Vijay Kumar and Camillo J. Taylor
\thanks{Shreyas S. Shivakumar, Neil Rodrigues, Alex Zhou, Ian D. Miller, Vijay Kumar and Camillo J. Taylor are with the GRASP Laboratory, School of Engineering and Applied Sciences,
        University of Pennsylvania, Philadelphia PA 19104
        {\tt\small \{sshreyas,rodri651, alexzhou, iandm, vijay,cjtaylor\}
        @seas.upenn.edu}}%
}

\begin{document}

\maketitle
\thispagestyle{empty}
\pagestyle{empty}

\begin{abstract}

In this work we propose long wave infrared (LWIR) imagery as a viable supporting modality for semantic segmentation using learning-based techniques. We first address the problem of RGB-thermal camera calibration by proposing a passive calibration target and procedure that is both portable and easy to use. Second, we present PST900, a dataset of 894 synchronized and calibrated RGB and Thermal image pairs with per pixel human annotations across four distinct classes from the DARPA Subterranean Challenge. Lastly, we propose a CNN architecture for fast semantic segmentation that combines both RGB and Thermal imagery in a way that leverages RGB imagery independently. We compare our method against the state-of-the-art and show that our method outperforms them in our dataset. 

\end{abstract}

\section{INTRODUCTION}

The ability to parse raw imagery and ascertain pixel-wise and region-wise semantic information is desirable for environment perception enabling advanced robot autonomy. Semantic segmentation is a major subject of robotics research and has applications ranging from medicine \cite{ronneberger2015u} and agriculture \cite{chen2017counting} to autonomous vehicles. Most popularly, convolutional neural networks (CNNs) have been applied to image classification tasks, where they dramatically out-perform their classical counterparts. CNNs have also grown in popularity as being highly effective for extracting semantic information from color images. The recent growth in autonomous vehicle research has driven the design of datasets, benchmarks and network architectures that focus on semantic segmentation from RGB imagery \cite{cordts2016cityscapes,zhou2017scene,alhaija2018augmented,lin2014microsoft,everingham2010pascal}.

Until recently, thermal cameras were primarily used by the military, with their cost and restricted usage making them difficult to acquire \cite{vollmer2017infrared}. However, over the last few years thermal cameras have become more easily accessible and their competitive prices have led to an increase in popularity. The primary use case thus far has been in surveillance, and most popular thermal cameras such as FLIR's range of LWIR cameras are specifically designed to identify human temperature signatures. There is a vast body of research in the field of thermal camera based human identification and tracking, which is out of the scope of our work.

We propose the usage of thermal cameras in addition to RGB cameras in challenging environments. Specifically, we look at environments with visibility and illumination limitations, such as in underground tunnels, mines and caves. We show that the additional information from the long-wave infrared spectrum can help to improve overall segmentation accuracy since it is not dependent on visible spectrum illumination which RGB cameras rely heavily upon. In this work, we also show that the segmentation of objects that do not possess very unique thermal signatures, such as hand-drills, also improves with the fusion of thermal information.

\begin{figure}[t]
\begin{center}
\includegraphics[width=0.95\linewidth]{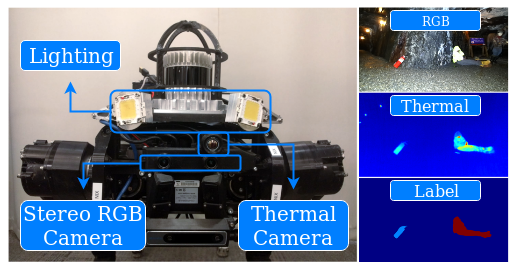}
\end{center}
   \caption{\textit{Left} - Quadruped mobile robot platform for data collection; sensor head includes a Stereolabs Zed Mini Stereo RGB camera, FLIR Boson 320 and an active illumintation setup. \textit{Right} - example RGB and Thermal imagery, with a human annotated segmentation label at the bottom.}
\label{fig:robot_data}
\end{figure}

Using thermal imagery in addition to RGB for general purpose semantic segmentation is a growing field of research with methods such as MFNet \cite{ha2017mfnet} and RTFNet \cite{sun2019rtfnet}, which are currently the latest and most popular CNN-based approaches. However, for these methods to generalize well and achieve state-of-the-art accuracy, they require large amounts of training data. Unlike for RGB imagery, large datasets of annotated thermal imagery for semantic segmentation are hard to find. Ha \etal present a dataset which is the most recent dataset for RGB and Thermal segmentation \cite{ha2017mfnet}. In our work, we present what we believe is the second dataset containing calibrated RGB and Thermal imagery that has per-pixel annotations across four different classes. 

We additionally propose a dual-stream CNN architecture that combines RGB and Thermal imagery for semantic segmentation. We design the RGB stream to be independently re-usable as it is easier to collect large amounts of RGB data and annotations. We design our Thermal stream to leverage the information learned from the RGB stream to refine and improve class predictions. In contrast to a single or tightly coupled network architecture like MFNet and RTFNet, we are able to leverage both modalities in a way that is able to achieve high accuracy while working in real-time on embedded GPU hardware. In summary, our contributions are as follows:

\begin{itemize}
  \item A method of RGB and LWIR (thermal) \textit{camera calibration} that uses no heated elements, allowing for faster, portable calibration in the field. 
  \item \textit{PST900 - Penn Subterranean Thermal 900} : A \textit{dataset} of approximately 900 annotated RGB and LWIR (thermal) images in both raw 16-bit and FLIR's AGC 8-bit format from a variety of challenging environments \cite{fliragc}. An additional 3416 annotated RGB images are also provided from these environments.
  \item A \textit{dual-stream CNN} architecture that is able to fuse RGB information with thermal information in such a way that allows for RGB stream re-usability and fast, real-time inference on embedded GPU platforms such as the NVIDIA Jetson TX2 and AGX Xavier.
  \item Extensive experiments comparing our method to similar approaches on both PST900 and the MFNet dataset.
\end{itemize}

\section{RELATED WORK}

There has been work in Thermal and RGB interaction in the form of cross modal prediction, where either RGB or Thermal imagery is used to predict the other \cite{kniaz2017thermalnet}. This idea of cross modal learning has also extended to stereo disparity estimation, where matching is done across the two different modalities \cite{treible2017cats,beaupre2019siamese}, resulting in the creation of interesting cross modal datasets such as LITIV \cite{torabi2012iterative} and St. Charles \cite{bilodeau2014thermal}. Since thermal cameras still operate at a lower resolution than similarly priced RGB cameras, Choi \etal and Feras \etal propose learning based enhancement \cite{choi2016thermal} and super-resolution methods \cite{almasri2018multimodal} for thermal imagery and use RGB imagery as a guide. \nocite{st2007combination}

\subsection{Semantic Segmentation with Thermal Images}

Qiao \etal propose a novel level set method for contour detection using an edge based active contour model designed specifically for thermal imagery \cite{qiao2017thermal}. 
Luo \etal use semantic information in an egocentric RGB-D-T SLAM pipeline \cite{luo2017scene} using a variant of YOLO on their RGB-D-T data \cite{redmon2016you}. Their experiments suggest that the model is most heavily benefited by the thermal modality and that thermal residues provide good indicators for action recognition tasks.

Directly relevant to our work is MFNet \cite{ha2017mfnet}, an RGB-T semantic segmentation network architecture proposed by Ha \etal. They present an RGBT dataset in urban scene settings for autonomous vehicles and a dual encoder architecture for RGB and Thermal image data. They show that this architecture performs better than naively introducing the thermal modality as an extra channel. Additionally, the authors state that with slightly misaligned RGB-T images, introducing thermal as a fourth channel can have detrimental effects to segmentation accuracy, often performing worse than RGB alone.

Recently, Sun \etal proposed a segmentation architecture that uses a dual ResNet encoder with a small decoder \cite{sun2019rtfnet}. The multimodal fusion is performed by an element-wise summation of feature blocks from both the RGB and Thermal encoder pathways. Their decoder architecture makes use of a novel Upception block which alternatingly preserves and increases spatial resolution while reducing channel count. They evaluate their network against popular semantic segmentation networks such as U-Net \cite{ronneberger2015u}, SegNet \cite{badrinarayanan2017segnet}, PSPNet \cite{zhao2017pyramid}, DUC-HDC \cite{wang2018understanding} and ERFNet \cite{romera2017erfnet}. They also compare their work against MFNet and show that RTFNet outperforms theirs on the MFNet dataset. From here on, we will compare our method to MFNet and RTFNet as they are the most relevant to our work.

\subsection{Calibration}

Our primary motivation for designing a calibration procedure was for portability and ease of use in resource constrained subterranean environments. Current calibration methods are either \textit{active:} where the calibration target is a thermal emitter or is externally heated to retain a thermal signature or \textit{passive:} where no explicit heat source is required. Popular among active methods is a paper calibration target heated by means of an external heat source such as a flood lamp in order to drive the temperature of the black ink higher than the white, resulting in an inverted checkerboard in the thermal camera. We initially used this technique and were able to calibrate RGB-T intrinsics and extrinsics successfully, but the process required significant effort to ensure that the amount of heat imparted to the checkerboard was sufficient to obtain sharp checkerboard corners consistently. This was specially difficult because of the fast cool-down time of the heated elements\cite{vidas2012mask}. Additionally, this required a large heat source (flood lamp), which was a burden to transport and use in the field. This motivated our search for a calibration target that was completely passive and could be used with off-the-shelf calibration tools.

Zoetgnande \etal propose an active method to calibrate a low-cost low-resolution stereo LWIR thermal camera system. They design a board with 36 bulbs and develop a sub-pixel corner estimation algorithm to detect these heat signatures against a wooden calibration frame \cite{zoetgnande2019robust}. Zalud \etal detail a five camera RGB-D(ToF)-T system for robotic telepresence \cite{zalud2013fusion}. Their calibration target was an actively heated aluminum checkerboard with squares cut out, placed in front of a black insulator. Rangel \etal present a detailed comparison of different calibration targets and methods for calibrating Depth and Thermal cameras together \cite{rangel20143d}. They compare active and passive methods and pick a calibration target with circular cutouts that requires minimal heating prior to calibration to appear in both depth and thermal imagery. 

Tarek \etal present a visual odometry method that uses stereo thermal cameras \cite{mouats2015thermal}. While this work is not directly relevant to ours, they propose an interesting \textit{passive} calibration method: they design a calibration board from highly polished aluminum with matte black squares applied to it. The calibration board is placed such that the aluminium reflects the \textit{cold sky}, leading to a high contrast between the aluminium and the squares. While this is a step in the right direction for portability and ease of use, it is still challenging for us to rely on the \textit{cold sky effect} in the event of calibration in subterranean environments. Our proposed method draws inspiration from this method to propose a \textit{passive} method that will work in both environments.  

\begin{figure}[t]
\begin{center}
\vspace{0.15cm}
\includegraphics[width=0.95\linewidth]{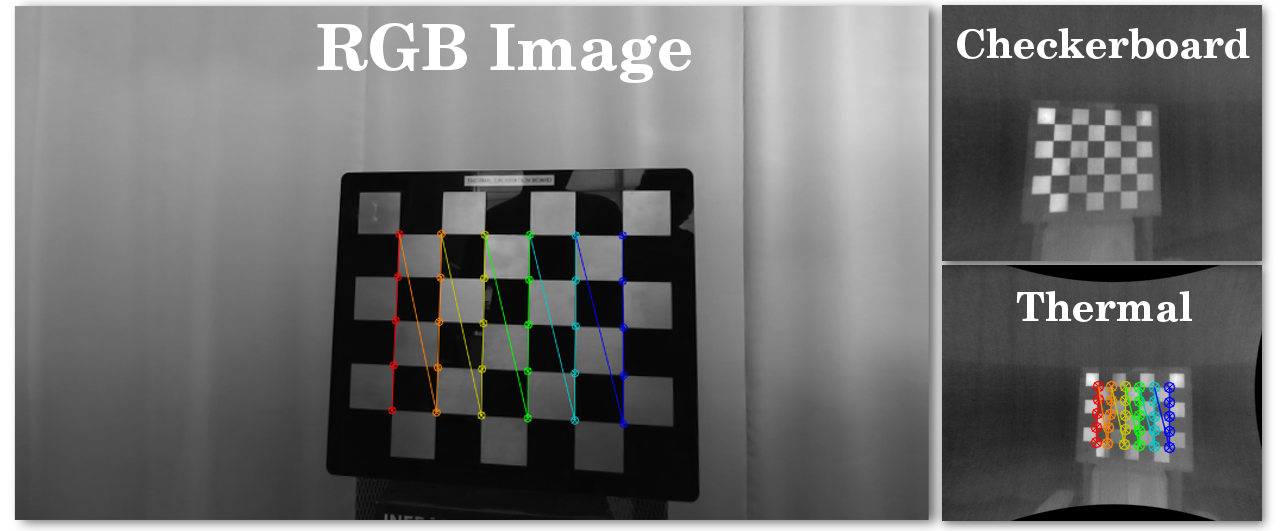}
\vspace{-0.25cm}
\end{center}
   \caption{Calibration: \textit{Left} - Undistorted rgb image with detected corners overlaid. \textit{Right} - Undistorted thermal image with detected corners overlaid, example thermal image containing our proposed calibration target; As this figure suggests, even with no heated elements, sufficient contrast can be achieved between the aluminum checkers and the backboard.}
\label{fig:rgbt_calibration}
\vspace{-0.45cm}
\end{figure}

\section{METHOD}

\subsection{RGB-D-T(LWIR) Calibration}

\subsubsection{Reflectivity Based Calibration}

A characteristic of LWIR reflections from metallic surfaces such as aluminum and copper is that detections in this band have much lower emissivities ($\epsilon$) compared to other bands such as medium wave infra red (MWIR) \cite{vollmer2017infrared}. This leads to strong detections of reflections $(1 - \epsilon)$ even if the material is rough and unpolished. We propose a calibration target that uses thermal reflectivity, specifically the reflections of the thermographer, i.e the person calibrating the system, to illuminate sand blasted aluminum squares mounted on a black acrylic background to form a checkerboard (Fig \ref{fig:rgbt_calibration}). We decided against polishing either surfaces since we found that highly polished surfaces tended to interfere with corner detection in RGB imagery. In practice the silver aluminum checkers appear to be at a higher temperature than the black acrylic background in the thermal imagery. We also achieve sufficient contrast in the RGB imagery to use existing checkerboard detectors. 
Note that since in both modalities, thermal and RGB, the silver checkers appear to have higher intensity values the correspondence between the two images is direct and no correspondence inversions were required such as those required when heating black ink on paper. For corner detection, OpenCV's chessboard detector performed poorly in both RGB and Thermal, and we used a C++ implementation of \textit{libcbdetect} instead \cite{geiger2012automatic}. Once the checkerboards are detected, we use OpenCV's fisheye camera calibration toolbox to first calibrate RGB and Thermal intrinsics followed by extrinsics. 

\subsubsection{Thermal-RGB Alignment}

\begin{figure}[t]
\begin{center}
\vspace{0.15cm}
\includegraphics[width=0.95\linewidth]{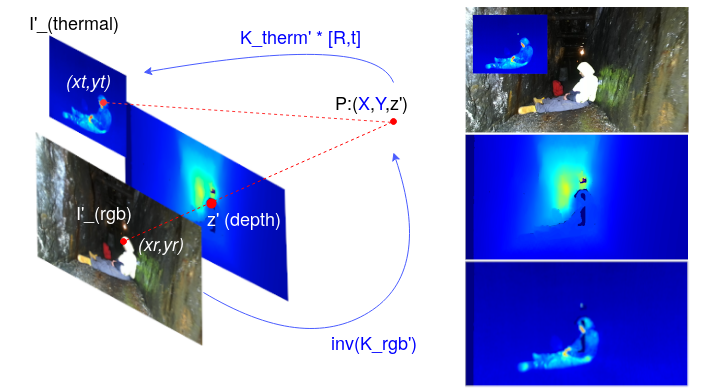}
\vspace{-0.25cm}
\end{center}
   \caption{Calibration for RGB-T Alignment: \textit{Left} - With intrinsics and extrinsics known, projecting thermal ($320\times256$) onto RGB ($1280\times720$) is achieved by projecting 2D co-ordinate locations in RGB into 3D using $K'_{rgb}$ and using the stereo depth for Z. These 3D co-ordinates are then projected onto the thermal frame using the calibrated $R$ and $t$ matrices along with $K'_{thermal}$ to obtain a mapping $I'_{therm} \mapsto I'_{rgb}$. \textit{Right} - Original undistorted RGB and thermal images, depth image from stereo and thermal projected onto RGB frame.}
\label{fig:rgbt_alignment}
\vspace{-0.25cm}
\end{figure}

Let $K_{rgb}$, $K_{thermal}$ and $D_{rgb}$, $D_{thermal}$ be the camera matrix and distortion co-efficients obtained from intrinsic calibration of the RGB camera and thermal camera respectively. From these parameters, we obtain $K'_{rgb}$ and $K'_{thermal}$, i.e the undistorted camera matrices for both cameras. The RGB camera is a Stereolabs Zed Mini and its intrinsics are modeled with a plumb-bob model, whereas the Thermal camera is a FLIR Boson 320 2.3mm and its intrinsics are captured with a fisheye model.

To register each pixel from the Thermal camera to the frame of the RGB camera, we require a mapping of all image co-ordinates $I'_{thermal}$ in the thermal image to image co-ordinates $I'_{rgb}$ in the rgb image. We achieve this by first projecting all RGB co-ordinates into 3D ($P$) using the depth image $D_{depth}$ acquired from stereo depth estimation. We then identify a mapping to the thermal frame by projecting these points back onto the thermal camera frame as seen in Fig \ref{fig:rgbt_alignment}. 

For some co-ordinate $i$, let $I'_{i,rgb}=(xr,yr,1)$ be a point in the undistorted RGB image. Let  $D_{i,depth}$ denote the depth provided by the stereo system at that pixel. This point is projected to 3D location $P_{i}$ using $(K')^{-1}_{i,rgb}$ as follows:
\begin{equation}
P_{i} = ((K'_{rgb})^{-1} \times (\{I'_{i,rgb}\},1))\times D_{i,depth} 
\end{equation}
This 3D co-ordinate $P_{i}=(X,Y,Z)$ can then be projected onto the thermal frame using the calibrated extrinsics $R^{thermal}_{rgb}$ and $t^{thermal}_{rgb}$,
\begin{equation}
I'_{i,thermal} = K'_{thermal} \times \lbrack R^{thermal}_{rgb} , t^{thermal}_{rgb} \rbrack \times P_{i} 
\end{equation}
where $I'_{i,thermal}=(xt,yt,1)$ is the point in the thermal image to which $I'_{i,rgb}$ is mapped to. We now have $I'_{i,rgb} \mapsto I'_{i,thermal}$ from which we find $I'_{i,thermal} \mapsto I'_{i,rgb}$. While calculating the inverse mapping, we handle the issue of parallax and the many-to-one mapping by choosing the closest 3D point during re-projection. In our dataset, we provide aligned thermal imagery with holes, but provide a simple interpolation script to perform hole filling.


\subsection{Penn Subterranean Thermal 900 Dataset (PST900):}

\begin{figure*}[t!]
\begin{center}
\vspace{0.15cm}
\includegraphics[width=0.132\linewidth]{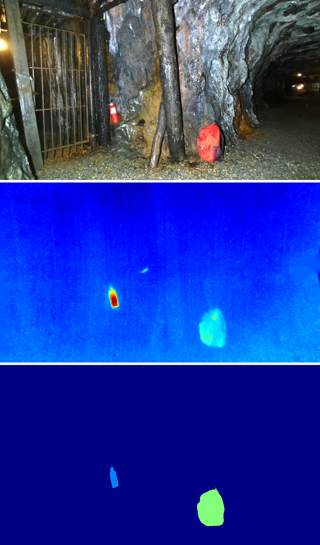}
\includegraphics[width=0.132\linewidth]{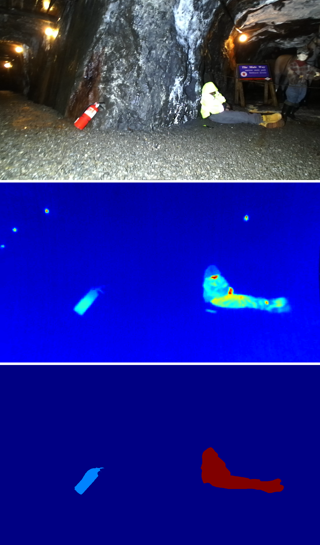}
\includegraphics[width=0.132\linewidth]{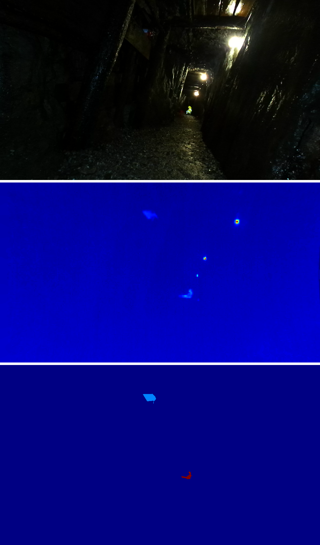}
\includegraphics[width=0.132\linewidth]{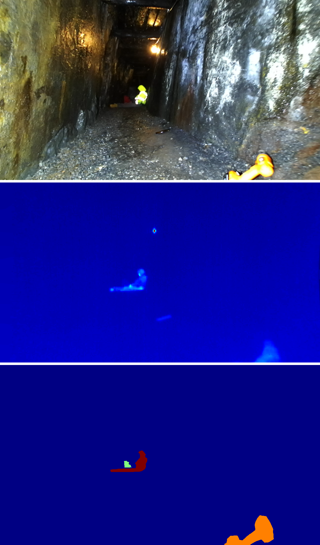}
\includegraphics[width=0.132\linewidth]{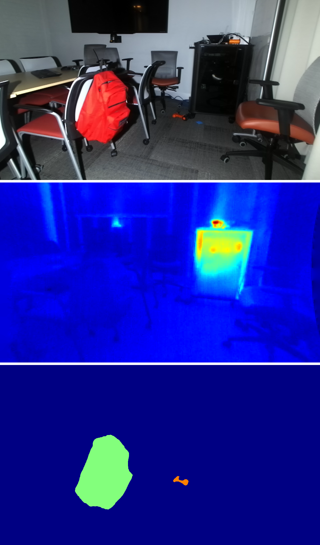}
\includegraphics[width=0.132\linewidth]{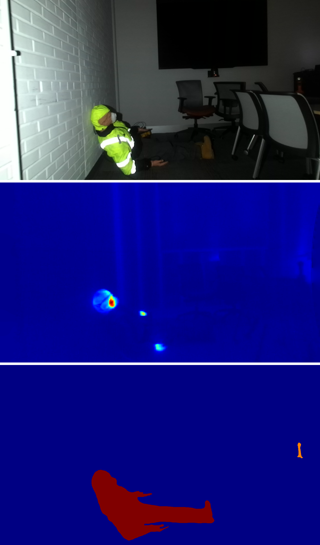}
\includegraphics[width=0.132\linewidth]{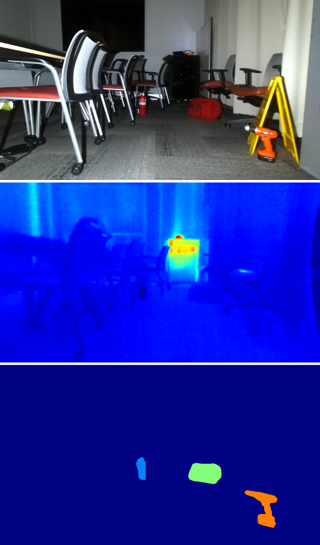}
\vspace{-0.25cm}
\end{center}
   \caption{PST900 Dataset: Examples from our proposed dataset; The \textit{first row} contains example RGB image data captured from the Stereolabs Zed Mini camera. The \textit{second row} contains aligned thermal imagery (visualized in 8-bit color) captured from the FLIR Boson camera. The \textit{last row} contains human annotated per-pixel labels for each of the four classes.}
\label{fig:rgbt_dataset}
\vspace{-0.45cm}
\end{figure*}

In this section, we present our dataset of 894 aligned pairs of RGB and Thermal images with per-pixel human annotations. This dataset was driven by the needs of the \textit{DARPA Subterranean Challenge}\footnote{https://www.subtchallenge.com/}, where a set of 4 visible artifacts (\textit{fire-extinguisher}, \textit{backpack}, \textit{hand-drill}, \textit{survivor : thermal mannequin, human}) are to be identified in challenging underground environments where there are no guarantees of environmental illumination or visibility. We therefore resort to equipping our robots and data collection platforms with high intensity LEDs. Our sensor head, as shown in Fig \ref{fig:robot_data}, consists of a Stereolabs ZED Mini stereo camera and a FLIR Boson 320 camera. We intentionally opted for a wider field of view thermal camera for a greater overlap between the two sensors. We design a calibration procedure to obtain camera intrinsics and system extrinsics as mentioned in the previous section. We then collect representative data from multiple different environments with varying degrees of lighting, these environments include the Number 9 Coal Mine in Lansford PA, cluttered indoor and outdoor spaces as seen in Fig \ref{fig:rgbt_dataset}. 

In addition to this corpus of labelled and calibrated RGB-Thermal data, we additionally provide a much larger set of similarly annotated RGB-only data, collected over a much larger set of environments. Our proposed method is able to leverage both datasets to produce efficient and accurate predictions.

Labels are acquired from a pool of human-annotators within our laboratory. The human-annotators are briefed on the different artifacts present, and the difficulty of visible identification of small artifacts such as the hand-drill. Annotations are made per-pixel and each set of {rgb, thermal and label} is verified by the authors for accuracy. Artifacts incorrectly labeled or missed artifacts are sent back for re-labeling and re-verification. With this process in place, we acquired our dataset of 894 aligned and annotated RGB-thermal image pairs and 3416 annotated RGB images. Our dataset is made publicly available to the community along with a basic toolkit here: \emph{https://github.com/ShreyasSkandanS/pst900\_thermal\_rgb}. 

\begin{table}
\centering
\caption{PST900 Dataset Class Imbalance: The first table shows the ratio of pixels of a particular class to the total pixels in the dataset. The second table shows the ratio of instances of a particular object to the total images in the dataset.}
\begin{tabular}{|c|c|c|c|c|c|}
\hline
\multicolumn{6}{|c|}{\textit{Class Imbalance} - per pixel (in \%)}\\
\hline
Dataset & \textit{bg} & \textit{fire-ext} & \textit{backpack} & \textit{drill} & \textit{survivor} \\
\hline
RGBDT & 96.9845 & 0.2829 & 1.1962 & 0.1764 & 1.3597 \\
RGB & 98.5003 & 0.1835 & 0.5090 & 0.0941 & 0.7129  \\
\hline
\multicolumn{6}{|c|}{\textit{Class Imbalance} - per instance (in \%)}\\
\hline
Dataset & \textit{bg} & \textit{fire-ext} & \textit{backpack} & \textit{drill} & \textit{survivor} \\
\hline
RGBDT & 100.0 & 28.6353 & 47.0917 & 34.0044 & 38.7024 \\
RGB & 100.0 & 23.9842 & 29.0225 & 20.1996 & 22.0571 \\
\hline
\end{tabular}
\label{tab:class_balance}
\vspace{-0.45cm}
\end{table}

\subsection{Segmentation}

\begin{figure}[t]
\begin{center}
\includegraphics[width=235px]{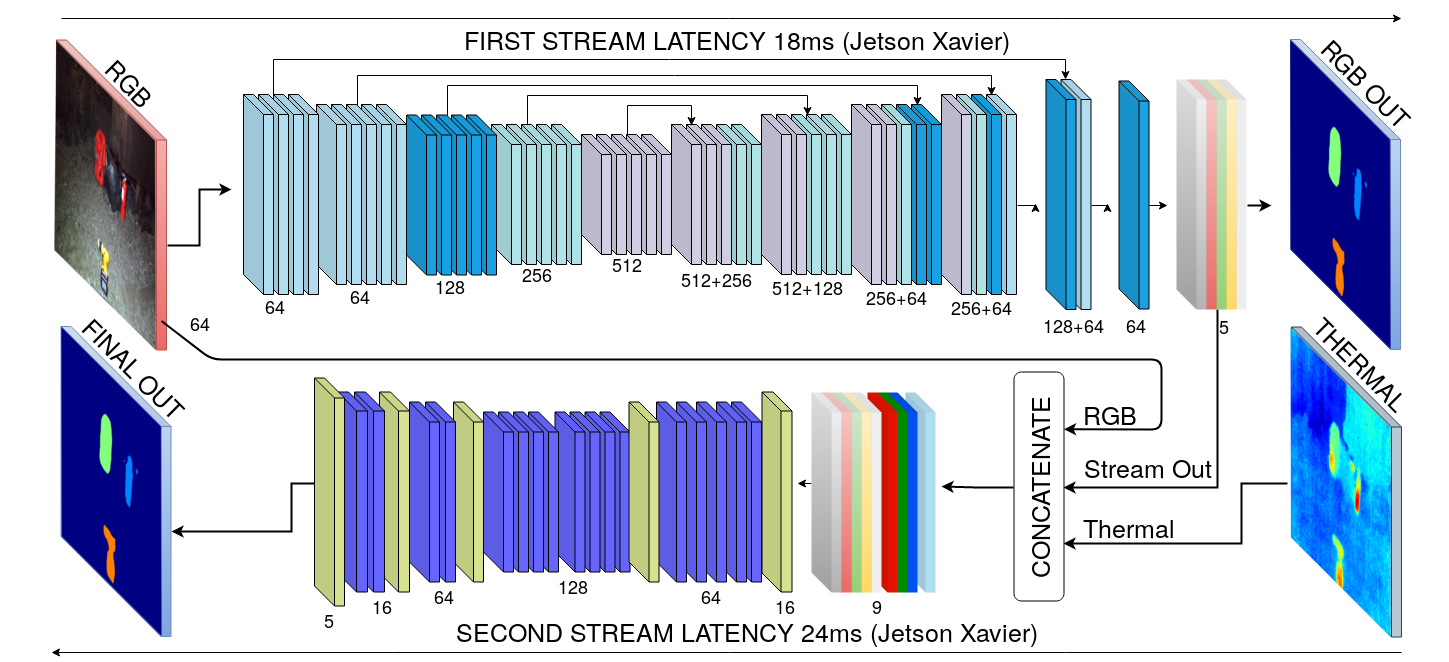}
\end{center}
   \caption{Our proposed network architecture that is essentially two streams, a single independent RGB based network that is based on ResNet-18 and UNet, and a second stream that depends on the output of the first network as well as thermal.}
\label{fig:pst_network}
\vspace{-0.45cm}
\end{figure}

Collecting large amounts of RGB data and acquiring accurate per-pixel human annotations is significantly easier and cheaper than collecting calibrated and aligned RGB-T data. Therefore, we designed a network that leverages this fact by having an independent RGB stream that can be trained without thermal data. We introduce the thermal modality to the output of this stream to further improve the initial results. We propose a sequential, dual stream architecture that draws influence from ResNet-18 \cite{he2016deep}, UNet \cite{ronneberger2015u} and ERFNet \cite{romera2017erfnet} and show that our design is \textit{efficient} allowing real-time inference on embedded hardware, \textit{flexible} since the early exit provides a coarser prediction quickly, and \textit{accurate} outperforming other methods on our dataset and showing competitive performance on the MFNet dataset.

The RGB stream is a ResNet-18 architecture with an encoder-decoder skip-connection scheme similar to UNet. Our network heavily based on the implementation of Usuyama \etal \cite{usuyama} and is shown in Fig \ref{fig:pst_network}. The network is first trained on annotated RGB images only. We use a weighted negative log-likelihood loss during training and select the model with the highest mean Intersection-over-Union (mIoU). The weights for the loss function are calculated using the weighting scheme proposed by Paszke \etal \cite{paszke2016enet}. On datasets such as ours, a weighting scheme is necessary given the dramatic imbalance between background and foreground classes as seen in Table \ref{tab:class_balance}. Once the network is trained, we remove the final softmax operator which results in what is intuitively a per-pixel confidence volume for the different classes in our dataset. We use this volume, along with the thermal modality, as input to our next fusion stream.    

The Fusion stream takes as it's input the confidence volume along with the input thermal imagery and color image. The information is concatenated and passed to an ERFNet-based encoder-decoder architecture \cite{romera2017erfnet}. Our architecture differs in that it has a larger set of initial feature layers to account for the larger input. Additionally, we use fewer layers at the end of the encoder. We then freeze the RGB stream and train this entire architecture as a whole using the same loss function as before. Once again, we select our best model to be the one with the lowest mean IoU value.

\begin{table*}[htb!]
\centering
\caption{Results on MFNet dataset: We compare our method to six different semantic segmentation networks. Aside from MFNet and RTFNet, the RGB-T mode indicates a naive fusion of thermal by adding it as a fourth channel to the network input.}
\vspace{-0.25cm}
\begin{tabular}{|c|c|c|c|c|c|c|c|c|c|c|c|c|}
 \hline
 \multicolumn{12}{|c|}{Dataset: \textbf{MFNet Dataset}} \\
 \hline
 Network & Mode & \textit{Background} & \textit{Car} & \textit{Person} & \textit{Bike} & \textit{Curve} & \textit{Car Stop} & \textit{Guardrail} & \textit{Color Cone} & \textit{Bump} & \textbf{mIoU}\\
 \hline
 ERFNet & RGB  & 0.9589 & 0.7051 & 0.4176 & 0.5029 & 0.2627 & 0.2070 & 0.0638 & 0.3643 & 0.3069 & 0.4199\\
 ERFNet & RGB-T & 0.9694 & 0.7384 & 0.6465 & 0.5128 & 0.3773 & 0.1833 & 0.0362 & \textbf{0.4023} & 0.4592 & 0.4806\\
 MAVNet & RGB & 0.8264 & 0.4156 & 0.1304 & 0.2878 & 0.0530 & 0.0290 & 0.0019 & 0.0560 & 0.0115 & 0.2014\\
 MAVNet & RGB-T & 0.8843 & 0.3754 & 0.3975 & 0.1521 & 0.0850 & 0.0240 & 0.0000 & 0.0400 & 0.0440 & 0.2226\\
 UNet* & RGB & 0.9620 & 0.6520 & 0.4260 & 0.4780 & 0.2780 & 0.2080 & 0.0000 & 0.3580 & 0.3100 & 0.4080\\
 UNet* & RGB-T & 0.9690 & 0.6620 & 0.6050 & 0.4620 & 0.4160 & 0.1790 & 0.0180 & 0.3060 & 0.4420 & 0.4510\\
 Fast-SCNN & RGB & 0.9477 & 0.4075 & 0.2334 & 0.2237 & 0.2358 & 0.0945 & 0.0166 & 0.2242 & 0.2165 & 0.2888\\
 Fast-SCNN & RGB-T & 0.9616 & 0.4263 & 0.5141 & 0.4735 & 0.2456 & 0.1087 & 0.0000 & 0.2230 & 0.3462 & 0.3283\\ 
 MFNet & RGB-T & 0.9635 & 0.6154 & 0.5530 & 0.4341 & 0.2231 & 0.0797 & 0.0028 & 0.2088 & 0.2471 & 0.3697\\
 RTFNet-50 & RGB-T  & 0.9789 & 0.8388 & 0.6676 & 0.6010 & \textbf{0.4328} & 0.1260 & 0.0551 & 0.2660 & \textbf{0.5684} & 0.5038\\
 RTFNet-152 & RGB-T & \textbf{0.9797} & \textbf{0.8710} & \textbf{0.6700} & \textbf{0.6100} & 0.4161 & 0.2360 & 0.0300 & 0.3200 & 0.4947 & \textbf{0.5141}\\
 \hline
 Ours: \textit{RGB Stream} & RGB & 0.9678 & 0.7673 & 0.4873 & 0.5532 & 0.2917 & \textbf{0.2785} & \textbf{0.1525} & 0.3580 & 0.4264 & 0.4776\\
 Ours: \textit{Na\"ive} & RGB-T & 0.9690 & 0.7298 & 0.6200 & 0.5166 & 0.3872 & 0.2011 & 0.0150 & 0.3691 & 0.4421 & 0.4700\\
 Ours: \textit{Full} & RGB-T & 0.9701 & 0.7684 & 0.5257 & 0.5529 & 0.2957 & 0.2509 & 0.151 & 0.3936 & 0.4498 & 0.4842\\
 \hline
\end{tabular}
\label{tab:mfnet_data}
\end{table*}

\begin{table*}[t]
\centering
\caption{Results on PST900 Dataset: Aside from MFNet and RTFNet, the RGB-T mode indicates a naive fusion of thermal by adding it as a fourth channel to the network input. We also list inference latency (in ms) on an NVIDIA AGX Xavier.}
\vspace{-0.25cm}
\begin{tabular}{|c|c|c|c|c|c|c|c|c|c|}
 \hline
 \multicolumn{9}{|c|}{Dataset: \textbf{PST900 Dataset}} \\
 \hline
 Network & Mode & \textit{Background} & \textit{Fire-Extinguisher} & \textit{Backpack} & \textit{Hand-Drill} & \textit{Survivor} & \textbf{mIoU} & \textbf{ms} (Xavier)\\
 \hline
 ERFNet & RGB & 0.9869 & 0.6118 & 0.6528 & 0.4240 & 0.4169 & 0.6185 & 30\\
 ERFNet & RGB-T & 0.9873 & 0.5879 & 0.6808 & 0.5276 & 0.3438 & 0.6255 & 31\\
 MAVNet & RGB & 0.9822 & 0.2831 & 0.5850 & 0.3367 & 0.0901 & 0.4551 & 16\\
 MAVNet & RGB-T & 0.9789 & 0.2258 & 0.5152 & 0.3194 & 0.3473 & 0.4774 & 17\\
 UNet & RGB & 0.9843 & 0.4928 & 0.6364 & 0.4026 & 0.2337 & 0.5499 & 12\\
 UNet & RGB-T & 0.9795 & 0.4296 & 0.5289 & 0.3827 & 0.3164 & 0.5274 & 12\\
 Fast-SCNN & RGB & 0.9857 & 0.3454 & 0.6679 & 0.2063 & 0.2053 & 0.4822 & 18\\
 Fast-SCNN & RGB-T & 0.9851 & 0.3548 & 0.6460 & 0.1550 & 0.2168 & 0.4715 & 18\\
 MFNet & RGB-T & 0.9863 & 0.6035 & 0.6427 & 0.4113 & 0.2070 & 0.5702 & 23\\
 RTFNet-50 & RGB-T & 0.9884 & 0.4349 & 0.7058 & 0.0100 & 0.2800 & 0.4840 & 42\\ 
 RTFNet-152 & RGB-T & \textbf{0.9892} & 0.5203 & \textbf{0.7530} & 0.2537 & 0.3643 & 0.5761 & 127\\ 
 \hline
 Ours: \textit{RGB Stream} & RGB & 0.9883 & 0.6814 & 0.6990 & 0.5151 & 0.4989 & 0.6765 & 18\\
 Ours: \textit{Na\"ive} & RGB-T & 0.9852 & 0.6253 & 0.5961 & 0.4708 & 0.4339 & 0.6223 & 20\\
 Ours: \textit{Full} & RGB-T & 0.9885 & \textbf{0.7012} & 0.6920 & \textbf{0.5360} & \textbf{0.5003} & \textbf{0.6836} & 42\\
 \hline
\end{tabular}
\label{tab:penn_data}
\end{table*}

\section{RESULTS AND ANALYSIS}

In this section, we compare our method against relevant methods MFNet \cite{ha2017mfnet}, and RTFNet \cite{sun2019rtfnet}. We also compare our method against na\"ive RGB-T fusion implementations on relevant segmentation networks such as ERFNet \cite{romera2017erfnet}, MAVNet \cite{mavnet}, Fast-SCNN \cite{poudel2019fast} and UNet \cite{ronneberger2015u}. In our na\"ive implementations of RGB-T segmentation networks, we introduce the thermal modality by concatenating the thermal image as a fourth channel to the original RGB input. For all our experiments, we compare both RGB and RGB-T performance as seen in Table \ref{tab:mfnet_data} and Table \ref{tab:penn_data}. We measure performance using \textit{mean Intersection over Union (mIoU)} across all classes. We train all the models in PyTorch using an NVIDIA DGX-1. For MFNet and RTFNet, we use author-recommended batch sizes, loss functions and training scripts where applicable. For the rest of the networks, we use a fixed batch size, learning rate, and loss function across all experiments. We measure inference latency in milliseconds on an NVIDIA AGX Xavier embedded GPU device, which is the central compute unit on-board our mobile robot platforms. To allow for fair comparison between methods, we use PST900 RGBT data and exclude the RGB only data.

\subsection{MFNet Dataset}

As shown in Table \ref{tab:mfnet_data}, the best performing method is RTFNet, with our method following second. MAVNet achieves the lowest scores on this dataset, reaching a maximum mIoU of 22.26\% with na\"ive fusion of thermal information. Nguyen \etal design this network with performance as a primary objective \cite{mavnet}, which is supported by the low inference latency of this network, as seen in Table \ref{tab:penn_data}. Fast-SCNN achieves an mIoU of 28.88\% and 32.83\% with RGB and RGB-T modalities respectively. The largest increase in class IoU between these two models is seen in the \textit{Person} class, which comports with the intuition that humans have a strong unique thermal signature, and the network is able to narrow in on this to improve its overall accuracy. However, Fast-SCNN was originally designed for high resolution data, which could explain the low performance \cite{poudel2019fast}. We were unable to achieve the results mentioned by Sun \etal with UNet on this dataset and therefore for comparisons, we refer to their experiments since we use the same training and validation split. RTFNet performs better than other methods in both ResNet-50 and ResNet-152 variants. This is followed by our method, which closely outperforms ERFNet with a na\"ively added thermal fourth channel. Our network achieves 48.42\% mIoU on this dataset and from a performance perspective, is roughly 4 times faster than RTFNet-152 as shown in Table \ref{tab:penn_data}.

\subsection{PST900 Dataset}

\begin{figure*}[t]
\begin{center}
\includegraphics[width=470px]{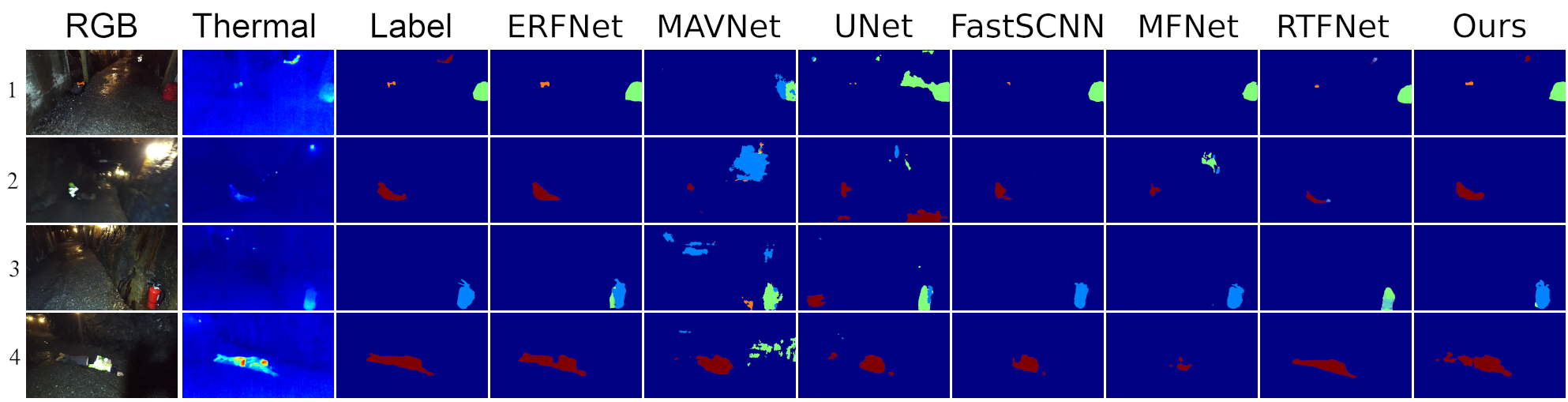}
\vspace{-0.25cm}
\end{center}
   \caption{\textit{Qualitative Results} on PST900: Our method is able to perform accurate segmentation of distant objects such as the \textit{Survivor} in \textit{(1)}. Our method also performs well in blurred and and visually degraded settings such as in \textit{(2)} where motion blur degrades the RGB image but the thermal image is able to help produce an accurate prediction. Example \textit{(3)} shows a clearly visible \textit{Fire Extinguisher} that is mistakenly identified by RTFNet-152, potentially because of weak thermal cues in the dataset and because of insufficient learning of color cues. Example \textit{(4)} shows our method perform poorly on the \textit{Survivor}; a portion of the survivor's leg appears to be near ground temperature and this leads to noisy segmentation in that region.}
\label{fig:qualcomp}
\vspace{-0.45cm}
\end{figure*}

All the above networks are trained and evaluated on our PST900 dataset. Our method achieves the best performance with 68.36\% mIoU and RTFNet-152 at 57.61\% mIoU; results are shown in Table \ref{tab:penn_data}. Qualitative comparisons shown in Fig \ref{fig:qualcomp}. The measured latency of our proposed method on our embedded GPU hardware is approximately 42ms, which is significantly faster than RTFNet-152. For the training and evaluation of the networks compared here, we use the originally recommended training parameters prescribed in their respective works. Aside from our method, the observed trend is similar to the previous experiments on the MFNet dataset, where RTFNet and ERFNet achieve high accuracy. However, RTFNet is outperformed by ERFNet on this dataset, achieving an mIoU of 62.55\% with na\"ive thermal fusion. Interesting to note is the na\"ive introduction of thermal information to UNet and Fast-SCNN results in lower performance than RGB alone, whereas this is not observed on the MFNet dataset. 

We posit that our dataset is significantly more challenging for thermal fusion networks since there is plenty of information available in RGB alone, making it potentially difficult to learn an informative correlation between both modalities. Additionally, our dataset contains the same object in situations where it is both above and below the ambient air temperature, resulting in an inversion in thermal imagery. This could be potentially challenging when learning RGB-Thermal correlations. Our hypothesis is strengthened by the two cases where performance degrades with the na\"ive introduction of thermal. The IoU for the \textit{Survivor} class increases, while objects that might have stronger color cues than thermal cues perform worse, like the \textit{backpack} class. 

We are able to achieve very competitive results with RGB alone, and accuracy further improves when the Fusion stream is added. We observe that our RGB Stream performs poorly with a na\"ive fusion approach. This also supports our hypothesis that for the task of learning correlated representations for RGB and Thermal, a \textit{late fusion} strategy is highly beneficial as opposed to fusing both modalities early on in the network architecture. This effect is exacerbated by the fact that our dataset contains objects of interest that have very strongly identifiable cues from RGB alone, such as the \textit{red} backpack and \textit{orange} hand-drill. When learning to identify these objects, the network may be prone to more heavily weight these attributes in RGB, and neglect the more subtle cues to be learned from Thermal.   

\section{CONCLUSIONS}

In summary, this work explores Thermal (LWIR) as a viable supporting modality for general semantic segmentation in challenging environments. We propose an RGB and Thermal camera calibration technique that is both portable and easy to use. To further research in this field, we also present PST900, a collection of 894 aligned and annotated RGB and Thermal image pairs and make this publicly available to the community. We compare various existing methods on this dataset and propose a dual-stream CNN architecture for RGB and Thermal guided semantic segmentation that achieves state of the art performance on this dataset. Our network works in real-time on embedded GPUs and can be use in mobile robotic systems. We also compare this method on the MFNet dataset and show that our method is competitive with existing methods. Additionally, we highlight the need for late fusion in these architectures by noting poor performance with na\"ive fusion approaches. We also discuss some of the challenges of RGB-Thermal fusion for object identification, such as when objects of interest may not have easily discernible thermal signatures but have strong cues from RGB. 

\balance 









\bibliographystyle{IEEEtran}
\bibliography{ref}

\end{document}